# Multivariate Information Bottleneck

**Nir Friedman    Ori Mosenzon    Noam Slonim    Naftali Tishby**
School of Computer Science & Engineering, Hebrew University, Jerusalem 91904, Israel
{nir, mosenzon, noamm, tishby }@cs.huji.ac.il

## Abstract

The *Information bottleneck method* is an unsupervised non-parametric data organization technique. Given a joint distribution $P(A, B)$, this method constructs a new variable $T$ that extracts partitions, or clusters, over the values of $A$ that are informative about $B$. The information bottleneck has already been applied to document classification, gene expression, neural code, and spectral analysis. In this paper, we introduce a general principled framework for multivariate extensions of the information bottleneck method. This allows us to consider multiple systems of data partitions that are inter-related. Our approach utilizes Bayesian networks for specifying the systems of clusters and what information each captures. We show that this construction provides insight about bottleneck variations and enables us to characterize solutions of these variations. We also present a general framework for iterative algorithms for constructing solutions, and apply it to several examples.

## 1 Introduction

Clustering, or data partitioning, is a common data analysis paradigm. A central question is understanding general underlying principles for clustering. One information theoretic approach to clustering is to require that clusters should capture only the "relevant" information in the data, where the relevance is explicitly determined by various components of the data itself. A common data type which calls for such a principle is co-occurrence data, such as verbs and direct objects in sentences [7], words and documents [1, 4, 11], tissues and gene expression patterns [14], galaxies and spectral components [10], etc. In most such cases the objects are discrete or categoric and no obvious "correct" measure of similarity exists between them. Thus, we would like to rely purely on the joint statistics of the co-occurrences and organize the data such that the "relevant information" among the variables is captured in the best possible way.

Formally, we can quantify the relevance of one variable, $A$, with respect to another one, $B$, in terms of the *mutual information*, $I(A; B)$. This well known quantity, defined as,

$$I(A; B) = \sum_{a,b} P(a, b) \log \frac{P(a, b)}{P(a)P(b)}$$

is symmetric, non-negative, and equals to zero if and only if the variables are independent. It measures how many bits are needed on the average to convey the information $A$ has about $B$ (or vice versa). The aim of information theoretic clustering is to find (soft) partitions of $A$'s values that are informative about $B$. This requires balancing two goals: we want to lose irrelevant distinctions made by $A$, and at the same time maintain relevant ones. A possible principle for extracting such partitions is the *information bottleneck* (IB) method [13]. Clustering is posed as a construction of a new variable $T$ that represents partitions of $A$. The principle is described by a variational tradeoff between the information we try to minimize, $I(A; T)$, and the one we try to maximize $I(T; B)$. We briefly review this principle and its consequences in the next section.

The main contribution of this paper is a general formulation of a *multivariate* extension of the information bottleneck principle. This extension allows us to consider cases where the clustering is relevant with respect to several variables, or where we construct several systems of clusters at the same time.

To give concrete motivation, we briefly mention two examples that we treat in detail in later sections. In *symmetric clustering* (also called *two-sided or double clustering*) we want to find two systems of clusters: one of $A$ and one of $B$ that are informative about each other. A possible application is relating documents to words, where we seek clustering of documents according to word usage, and a corresponding clustering of words. This procedure aims to find document clusters that correspond to different topics and at the same time identify cluster of words that characterize these topics [11]. Clearly, the two systems of clusters are in interaction, and we want a unifying principle that shows how to construct them simultaneously.

In *parallel clustering* we attempt to build several systems of clusters of the values of $A$. Our aim here is to capture independent aspects of the information $A$ conveys about $B$. A biological example is the analysis of gene expression data, where multiple independent distinctions about tissues (healthy vs. tumor, epithelial vs. muscle, etc.) are relevant for the expression of genes.

We present such tasks, and others, in our framework by specifying a pair of Bayesian networks. One network, $G_{in}$, represents which variables are compressed versions of the observed variables (each new variable compresses its parents in the network). The second network, $G_{out}$, represents which relations should be maintained or predicted (each variable is predicted by its parents in the network). We formulate the general principle as a tradeoff between the information each network carries. We want to minimize the information maintained by $G_{in}$ and to maximize the information maintained by $G_{out}$.

We further give another interpretation to this principle, as a tradeoff between compression of the source (given by $G_{in}$) and fitness to a *target model*, where the model is described by $G_{out}$. Using this interpretation we can think of our new principle as a generalized compression distortion tradeoff (as in rate-distortion theory [3]). This interpretation may allow us to investigate the principle in a general parametric setup. In addition, we show that, as with the original IB, the new principle provides us with self-consistent equations in the unknown probabilistic partition(s) which can be iteratively solved and shown to converge. We show how to combine this in a deterministic annealing procedure which enables us to explore the information tradeoff in an hierarchical manner. There are many possible applications for our new principle and algorithm. To mention just a few, we consider semantic clustering of words based on multiple parts of speech, complex gene-expression data analysis, and neural code analysis.

## 2 The Information Bottleneck

We start with some notation. We use capital letters, such as $A, B, T, X$, for random variable names and lowercase letters $a, b, t, x$ to denote specific values taken by those variables. Sets of variables are denoted by boldface capital letters $\mathbf{T}, \mathbf{X}$, and assignments of values to the variables in these sets are denoted by boldface lowercase letters $\mathbf{t}, \mathbf{x}$. The statement $P(a \mid b)$ is used as a shorthand for $P(A = a \mid B = b)$.

Tishby et al. [13] considered two variables, $A$ and $B$, with their (assumed given) joint distribution $P(A, B)$. Here $A$ is the variable we try to compress, with respect to the "relevant" variable $B$. Namely, we seek a (soft) partition of $A$ through an auxiliary variable $T$ and the probabilistic mapping $P(T \mid A)$, such that the the mutual information $I(A; T)$ is minimized (maximum compression) while the relevant information $I(T; B)$ is maximized. The dependency relations between the 3 variables can be described by the relations: $T$ independent of $B$ given $A$; and on the other hand we want to predict $B$ from $T$.

By introducing a positive Lagrange multiplier $\beta$, Tishby et al. formulate this tradeoff by minimizing the following Lagrangian,

$$\mathcal{L}[P(T \mid A)] = I(A; T) - \beta I(T; B)$$

where we take $P(A, B, T) = P(A, B) P(T \mid A)$.

By taking the variation (i.e derivative in the finite case) of $\mathcal{L}$ w.r.t. $P(T \mid A)$, under the proper normalization constraints, Tishby et al. show that the optimal partition satisfies,

$$P(t \mid a) = \frac{P(t)}{Z(a, \beta)} \exp\left(-\beta D(P(B \mid a) \| P(B \mid t))\right)$$

where $D(P \| Q) = E_P[\log \frac{P}{Q}]$ is the familiar KL divergence [3]. This equation must be satisfied self consistently.

The practical solution of these equations can be done by repeated iterations of the self-consistent equations, for every given value of $\beta$, similar to clustering by deterministic annealing [8]. The convergence of these iterations to a (generally local) optimum was proven in [13] as well.

## 3 Bayesian Networks and Multi-Information

A Bayesian network structure over a set of random variables $X_1, \ldots, X_n$ is a DAG $G$ in which vertices are annotated by names of random variables. For each variable $X_i$, we denote by $\mathbf{Pa}_{X_i}^G$ the (potentially empty) set of parents of $X_i$ in $G$. We say that a distribution $P$ is *consistent* with $G$, if $P$ can be factored in the form:

$$P(X_1, \ldots, X_n) = \prod_i P(X_i \mid \mathbf{Pa}_{X_i}^G)$$

and use the notation $P \models G$ to denote that.

One of the main issues that we will deal with is the amount of information that variables $X_1, \ldots, X_n$ contain about each other. A quantity that captures this is the *multi-information* given by

$$\begin{aligned} \mathcal{I}(X_1, \ldots, X_n) &= D(P(X_1, \ldots, X_n) \| P(X_1) \cdots P(X_n)) \\ &= E_P[\log \frac{P(X_1, \ldots, X_n)}{P(X_1) \cdots P(X_n)}]. \end{aligned}$$

The multi-information captures how close is the distribution $P(X_1, \ldots, X_n)$ to the factored distribution of the marginals. This is a natural generalization of the pairwise concept of mutual information. If this quantity is small, we do not lose much by approximating $P$ by the product distribution. Like mutual information, it measures the average number of bits that can be gained by a joint compression of the variables vs. independent compression.

When $P$ has additional known independence relations, we can rewrite the multi-information in terms of the dependencies among the variables:

**Proposition 3.1**: *Let $G$ be a Bayesian network structure over $\mathbf{X} = \{X_1, \ldots, X_n\}$, and let $P$ be a distribution over $\mathbf{X}$ such that $P \models G$. Then,*

$$\mathcal{I}(X_1, \ldots, X_n) = \sum_i I(X_i; \mathbf{Pa}_{X_i}^G).$$

That is, the multi-information is the sum of *local* mutual information terms between each variable and its parents. We denote the sum of these informations with respect to a network structure as:

$$\mathcal{I}^G = \sum_i I(X_i; \mathbf{Pa}_{X_i}^G).$$

When $P$ is not consistent with the DAG $G$, we often want to know how close is $P$ to a distribution that is consistent with $G$. That is, what is the distance (or *distortion*) of $P$ from its *projection* onto the sub-space of distributions consistent with $G$. We naturally define this distortion as

$$D(P\|G) = \min_{Q \models G} D(P\|Q) .$$

This measure has two immediate interpretations in terms of the graph $G$.

**Proposition 3.2**: *Let $G$ be a Bayesian network structure over $\mathbf{X} = \{X_1, \ldots, X_n\}$, and let $P$ be a distribution over $\mathbf{X}$. Assume that the order $X_1, \ldots, X_n$ is consistent with the DAG, then*

$$\begin{aligned} D(P\|G) &= \sum_i I(X_i; \{X_1, \ldots, X_{i-1}\} - \mathbf{Pa}^G_{X_i} \mid \mathbf{Pa}^G_{X_i}) \\ &= \mathcal{I}(X_1, \ldots, X_n) - \mathcal{I}^G \end{aligned}$$

Thus, we see that $D(P\|G)$ can be expressed as a sum of conditional information terms, where each term corresponds to a *Markov independence assumption* with respect to the order $X_1, \ldots, X_n$. Recall that the Markov independence assumptions (with respect to a given order) are necessary and sufficient to require the factored form of distributions consistent with $G$ [6]. We see that $D(P\|G)$ is measured in terms of the extent these independencies are violated, since $I(X_i; \{X_1, \ldots, X_{i-1}\} - \mathbf{Pa}^G_{X_i} \mid \mathbf{Pa}^G_{X_i}) = 0$ if and only if $X_i$ is independent of $\{X_1, \ldots, X_{i-1}\} - \mathbf{Pa}^G_{X_i}$ given its parents. Thus, $D(P\|G) = 0$ if and only if $P$ is consistent with $G$.

An alternative representation of this distance measure is given in terms of multi-informations, since we can think of $D(P\|G)$ as the amount of information between the variables that cannot be captured by the dependencies of the structure $G$.

## 4 Multi-Information Bottleneck Principle

The multi-information allows us to introduce a simple "lift-up" of the original IB variational principle to the multivariate case, using the semantics of Bayesian networks of the previous section. Given a set of observed variables, $\mathbf{X} = \{X_1, \ldots, X_n\}$, instead of one partition variable $T$, we now consider a set $\mathbf{T} = \{T_1, \ldots, T_k\}$, which correspond to different partitions of various subsets of the observed variables. More specifically, we want to "construct" new variables, where the relations between the observed variables and these new compression variables are specified using a DAG $G_{in}$ over $\mathbf{X} \cup \mathbf{T}$. Since we assume that the new variables in $\mathbf{T}$ are functions of the original variables, we restrict attentions to DAGs where the variables in $\mathbf{T}$ are leafs.[1] Thus, each $T_j$ is a stochastic function of a set of variables $\mathbf{Pa}^{G_{in}}_{T_j} \subseteq \mathbf{X}$. Once these are set, we have a joint distribution over the combined set of variables:

$$P(\mathbf{X}, \mathbf{T}) = P(\mathbf{X}) \prod_j P(T_j \mid \mathbf{Pa}^{G_{in}}_j) \qquad (1)$$

[1] It will be convenient to think of cases where $G_{in}$ restricted to $\mathbf{X}$ forms a complete graph. However, this is not crucial in the following development. To simplify the technical details, we do assume that $P(\mathbf{X})$ is consistent with $G_{in}$.

Analogously to the original IB formulation, the information that we would like to minimize is now given by $\mathcal{I}^{G_{in}}$. Minimizing this quantity attempts to make variables as independent of each other as possible. (Note that since we only modify conditional distributions of variables in $\mathbf{T}$, we cannot modify the dependencies among the original variables.)

The "relevant" information that we want to preserve is specified by another DAG, $G_{out}$. This graph specifies, for each $T_j$ which variables it predicts. These are simply its children in $G_{out}$. Conversely, we want to predict each $X_i$ (or $T_j$) by its parents in $G_{out}$. Thus, we think of $\mathcal{I}^{G_{out}}$ as a measure of how much information the variables in $\mathbf{T}$ maintain about their target variables. This suggest that we wish to maximize is $\mathcal{I}^{G_{out}}$.

The generalized Lagrangian can be written as

$$\mathcal{L}^{(1)}[P(T_1 \mid \mathbf{Pa}^{G_{in}}_{T_1}), \ldots, P(T_k \mid \mathbf{Pa}^{G_{in}}_{T_k})] = \mathcal{I}^{G_{in}} - \beta \mathcal{I}^{G_{out}},$$
(2)

and the variation is done subject to the normalization constraints on the partition distributions. It leads to tractable self-consistent equations, as we henceforth show.

It is easy to see that the form of this Lagrangian is a direct generalization of the original IB principle. Again, we try to balance between the information $\mathbf{T}$ loses about $\mathbf{X}$ in $G_{in}$ and the information it preserves with respect to $G_{out}$.

**Example 4.1**: As a simple example, consider application of the variational principle with $G_{in}$ and $G_{out}^{(a)}$ of Figure 1. $G_{in}$ specifies that $T$ compresses $A$ and $G_{out}^{(a)}$ specifies that we want $T$ to predict $B$. For this choice of DAGs, $\mathcal{I}^{G_{in}} = I(T; A) + I(A; B)$ and $\mathcal{I}^{G_{out}} = I(T; A)$. The resulting Lagrangian is

$$\mathcal{L}^{(1)} = I(T; A) + I(A; B) - \beta I(T; B)$$

Since, $I(A; B)$ is constant, we can ignore it, and we end up with a Lagrangian equivalent to that of the original IB method.

## 5 Analogous Variational Principle

We now describe a closely related variational principle. This one is based on approximating distributions by a class defined by the Bayesian network $G_{out}$, rather than on preservation of multi-information.

We face the problem of choosing the conditional distributions $P(T_j \mid \mathbf{Pa}^{G_{in}}_{T_j})$. Thus, we also need to specify what is exactly our target in constructing these variables. As with the original IB method, we are going to assume that there are two goals.

On the one hand, we want to compress, or partition, the original variables. As before the natural multivariate form of this is to minimize the multi-information of $P$. Since $P$ is consistent with $G_{in}$, we can denote this multi-information as $\mathcal{I}^{G_{in}}$.

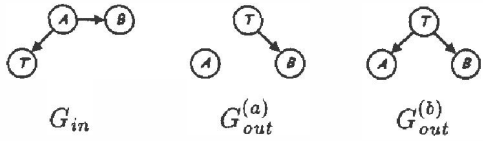

Figure 1: The source and target networks for the original IB.

While in the previous section the second goal was to preserve the multi-information to other (target, relevant) variables, here we think of a *target class* of model distributions, specified by a target Bayesian network. In that interpretation the compressed variables should help us in describing the joint distribution in a different desired structure. We specify this structure by another DAG $G_{out}$ that represents which independencies we would like to impose.

To make this more concrete consider the simple two-variable case shown in Figure 1. In this example, we are given the distribution of two variables $A$ and $B$. The DAG $G_{in}$ specifies that $T$ is a compressed version of $A$. On the other hand, we would like this $T$ to make $A$ and $B$ as independent as possible. A way of formally specifying this desire, is to specify the DAG $G_{out}^{(b)}$ of Figure 1. In this DAG, $T$ separates between $A$ and $B$.

The question now is how to force the construction of $P(T \mid A)$ such that it will lead to the independencies that are specified in the target DAG. Notice that these two DAGs are, in general, incompatible: Except for trivial cases, we cannot achieve both sets of independencies simultaneously. Instead, we aim to come as close as possible to achieving this by a tradeoff between the two. We formalize this by requiring that $P$ can be closely *approximated* by a distribution consistent with $G_{out}$. As previously discussed, a natural information theoretic measure for this approximation is $D(P\|G)$, the minimal KL divergence from $P$ to distributions consistent with $G_{out}$.

As before, we introduce a Lagrange multiplier that controls the tradeoff between these objectives. To distinguish it from the parameter we use above, we denote this parameter by $\gamma$. The Lagrangian we want to minimize in this formulation is thus:

$$\mathcal{L}^{(2)} = \mathcal{I}^{G_{in}} + \gamma D(P\|G_{out}) \qquad (3)$$

where the parameters that we can change during the minimization are again over the conditional distributions $P(T_j \mid \mathbf{Pa}_{T_j}^{G_{in}})$. The range of $\gamma$ is between 0, in which case we have a trivial solution in which the $T_j$'s are independent of their parents, and $\infty$, in which we strive to make $P$ as close as possible to $G_{out}$.

**Example 5.1:** Consider again the example of Figure 1 with $G_{in}$ and $G_{out}^{(b)}$. In this case, we have that $\mathcal{I}^{G_{in}} = I(A;B) + I(T;A)$ and $\mathcal{I}^{G_{out}} = I(T;A) + I(T;B)$. Using Proposition 3.2, we have that $D(P\|G_{out}) = \mathcal{I}^{G_{in}} - \mathcal{I}^{G_{out}}$. Putting, these together, we get the Lagrangian

$$\mathcal{L}^{(2)} = I(T;A) - \gamma I(T;B) + (1+\gamma)I(A;B)$$

Since, $I(A;B)$ is constant, we can ignore it, and we end up with a Lagrangian equivalent to that of the original IB method (setting $\gamma = \beta$). Thus, we can think of the IB as finding a compression $T$ of $A$ that results in a joint distribution that is as close as possible to the DAG where $A$ and $B$ are independent given $T$.

Going back to the general case, we can apply Proposition 3.2 to rewrite the Lagrangian in terms of multi-informations:

$$\begin{aligned}\mathcal{L}^{(2)} &= \mathcal{I}^{G_{in}} + \gamma(\mathcal{I}^{G_{in}} - \mathcal{I}^{G_{out}}) \\ &= (1+\gamma)\mathcal{I}^{G_{in}} - \gamma \mathcal{I}^{G_{out}}\end{aligned}$$

which is equivalent to the Lagrangian $\mathcal{L}^{(1)}$ presented in the previous section, under the transformation $\beta = \frac{\gamma}{1+\gamma}$. Where the range $\gamma \in [0,\infty)$ corresponds to the range $\beta \in [0,1)$. (Note that when $\beta = 1$, we have that $\mathcal{L}^{(1)} = D(P\|G_{out})$, which is the extreme case of $\mathcal{L}^{(2)}$.) Thus, from a mathematical perspective, $\mathcal{L}^{(2)}$ is a special case of $\mathcal{L}^{(1)}$ with the restriction $\beta \leq 1$.

This transformation raises the question of the relation between the two variational principles. As we have seen in Examples 4.1 and 5.1, we need different versions of $G_{out}$ in the two Lagrangians to reconstruct the original IB. To better understand the differences between the two, we consider the range of solutions for extreme values of $\beta$ and $\gamma$.

When $\beta \to 0$ and $\gamma \to 0$, both Lagrangians minimize the term $\mathcal{I}^{G_{in}}$. That is, the emphasis is on loosing information in the transformation from $\mathbf{X}$ to $\mathbf{T}$.

In the other extreme case, the two Lagrangians differ. When $\beta \to \infty$, minimizing $\mathcal{L}^{(1)}$ is equivalent to maximizing $\mathcal{I}^{G_{out}}$. That is, the emphasis is on preserving information about variables that have parents in $G_{out}$. For example, in the application of $\mathcal{L}^{(1)}$ in Example 4.1 with $G_{out}^{(a)}$, this extreme case results in maximization of $I(T;B)$. On the other hand, if we apply $\mathcal{L}^{(1)}$ with $G_{out}^{(b)}$, then we maximize $I(T;A) + I(T;B)$. In this case, when $\beta$ approaches $\infty$ information about $A$ will be preserved even if it is irrelevant to $B$.

When $\gamma \to \infty$, minimizing $\mathcal{L}^{(2)}$ is equivalent to minimizing $D(P\|G_{out})$. By Proposition 3.2 this is equivalent to minimizing the violations of conditional independencies implied by $G_{out}$. Thus, for $G_{out}^{(b)}$, this minimizes $I(A;B \mid T)$. Using the structure of $P$, we can write $I(A;B \mid T) = I(A;B) - I(T;B)$ (as implied by Proposition 3.2), and so this is equivalent to maximizing $I(T;B)$. If instead we use $G_{out}^{(a)}$, we minimize the information $I(A;B,T) = I(A;B) + I(T;A) - I(T;B)$. Thus, we minimize $I(T;A)$ while maximizing $I(T;B)$. Unlike, the application of $\mathcal{L}^{(1)}$ to $G_{out}^{(a)}$, we cannot ignore the term $I(A;T)$.

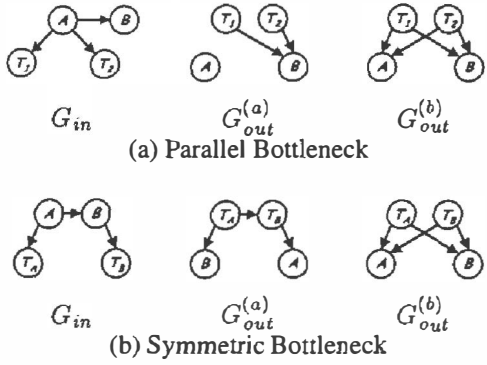

(a) Parallel Bottleneck

(b) Symmetric Bottleneck

Figure 2: The source and target networks for the parallel and symmetric Bottleneck examples.

To summarize, we might loosely say that $\mathcal{L}^{(1)}$ focuses on the edges that are present in $G_{out}$, while $\mathcal{L}^{(2)}$ focuses on the edges that are not present in $G_{out}$ (or more precisely, the conditional independencies they imply). This explains the somewhat different intuitions that apply to understanding the solutions found by the variational principles. Thus, although both variants can be applied to any choice of $G_{out}$, some choices might make more sense for $\mathcal{L}^{(1)}$ than for $\mathcal{L}^{(2)}$, and vice versa.

## 6 Bottleneck Variations

We now consider two examples of these principles applied to different situations.

**Example 6.1:** We first consider a simple extension of the original IB. Suppose, we introduce two variables $T_1$ and $T_2$. As specified in $G_{in}$ of Figure 2(a), both of these variables are stochastic functions of $A$. In addition, similarly to the original IB, we want $T_1$ and $T_2$ to extract information about $B$ from $A$. We call this example the *parallel* bottleneck, as $T_1$ and $T_2$ compress $A$ in "parallel".

The DAG $G_{out}^{(a)}$ specifies that $T_1$ and $T_2$ should predict $B$. Based on these two choices, $\mathcal{I}^{G_{in}} = I(A;B) + I(T_2;A) + I(T_1;A)$ and $\mathcal{I}_a^{G_{out}} = I(T_1,T_2;B)$. After dropping the constant term $I(A;B)$, the Lagrangian $\mathcal{L}^{(1)}$ can be written as

$$\mathcal{L}^{(1)} = I(T_1;A) + I(T_2;A) - \beta(I(T_1,T_2;B)) . \quad (4)$$

Thus, we attempt to minimize the information between $A$ and $T_1$ and $T_2$ while maximizing the information they preserve about $B$. Since $T_1$ and $T_2$ are independent given $A$, we can also rewrite[2]

$$I(T_1,T2;A) = I(T_1;A) + I(T_2;A) - I(T_1;T_2) . \quad (5)$$

Thus, minimizing $I(T_1;A) + I(T_2;A)$ is equivalent for minimizing $I(T1,T_2;A) + I(T_1;T_2)$. In other words, an-other interpretation for the above optimization is that we aim to find $T_1$ and $T_2$ that *together* try to compress $A$, preserve the information about $B$ and remain independent of each other as possible. In this sense, we can say that we are trying to decompose the information $A$ contains about $B$ into two "orthogonal" components.

---

[2]Proof: $I(T_1,T_2;A) = E[\log \frac{P(T_1,T_2,A)}{P(A)P(T_1,T_2)}]$
$= E[\log \frac{P(A)P(T_1|A)P(T_2|A)}{P(A)P(T_1,T_2)} \cdot \frac{P(T_1)P(T_2)}{P(T_1)P(T_2)}]$
$= I(T_1;A) + I(T_2;A) - I(T_1;T_2)$.

Recall, that using $\mathcal{L}^{(2)}$ we aim at minimizing violation of independencies in $G_{out}$. This suggests that the DAG $G_{out}^{(b)}$ of Figure 2(a) captures our intuitions above. In this DAG, $A$ and $B$ are independent given $T_1$ and $T_2$. Moreover, here again $G_{out}^{(b)}$ specifies an additional independence requirement over $T_1$ and $T_2$. To see that we examine the Lagrangian defined by the principle. In this case, $\mathcal{I}_b^{G_{out}} = I(T_1,T_2;A) + I(T_1,T_2;B)$. Using Eq.(5) and dropping the constant term $I(A;B)$, the Lagrangian $\mathcal{L}^{(2)}$ can be written as

$$\mathcal{L}^{(2)} = I(T_1;A) + I(T_2;A) + \gamma(I(T_1;T_2) - I(T_1,T_2;B)) .$$

Thus, again, we attempt to minimize the information between $T_1$ and $T_2$ while maximizing the information they *together* contain about $B$.

**Example 6.2:** We now consider the *symmetric* bottleneck. In this case, we want to compress $A$ into $T_A$ and $B$ into $T_B$ so that $T_A$ extracts the information $A$ contains about $B$, and at the same time $T_B$ extracts the information $B$ contains about $A$. The DAG $G_{in}$ of Figure 2(b) captures the form of the compression. The choice of $G_{out}$ is less obvious.

One alternative, shown as $G_{out}^{(a)}$ in Figure 2(b), attempts to make each of $T_A$ and $T_B$ sufficient to separate $A$ from $B$. As we can see, in this network $A$ is independent of $B$ and $T_B$ given $T_A$. Similarly, $T_B$ separates $B$ from the other variables. The structure of the network states that $T_A$ and $T_B$ are dependent of each other. Developing the Lagrangian defined by this network, we get:

$$\mathcal{L}_a^{(2)} = I(T_A;A) + I(T_B;B) - \gamma I(T_A;T_B)$$

Thus, on one hand we attempt to compress, and on the other hand we attempt to make $T_A$ and $T_B$ as informative about each other as possible. (Note that if $T_A$ is informative about $T_B$, then it is also informative about $B$.)

Alternatively, we might argue that $T_A$ and $T_B$ should each compress different "aspects" of the connection between $A$ and $B$. This intuition is specified by the target network $G_{out}^{(b)}$ of Figure 2(b). In this network $T_A$ and $T_B$ are independent of each other, and both are needed to make $A$ and $B$ conditionally independent. In this sense, our aim is to find $T_A$ and $T_B$ that capture independent attributes of the connection between $A$ and $B$. Indeed, following arithmetic similar to that of Example 6.1, we can write the Lagrangian as:

$$\mathcal{L}_b^{(2)} = I(T_A;A) + I(T_B;B) + \\ \gamma(2I(T_A;T_B) - I(T_B;A) - I(T_A;B))$$

That is, we attempt to maximize the information $T_A$ maintains about $B$ and $T_B$ about $A$, and at the same time try to minimize the information between $T_A$ and $T_B$.

## 7 Characterization of the Solution

In the previous sections we stated a variational principle. In this section we consider the form of the solutions of the principle. More precisely, we assume that $G_{in}$, $G_{out}$, and $\beta$ (or $\gamma$) are given. We now want to describe the properties of the distributions $P(T_j \mid \mathbf{Pa}^{G_{in}}_{T_j})$. We present this characterization for the Lagrangians of the form of $\mathcal{L}^{(1)}$. However, we can easily recover the corresponding characterization for Lagrangians of the form $\mathcal{L}^{(2)}$ (using the transformation $\beta = \frac{\gamma}{1+\gamma}$).

In the presentation of this characterization, we need some additional notational shorthands. We denote by $\mathbf{U}_j = \mathbf{Pa}^{G_{in}}_{T_j}$, $\mathbf{V}_{T_j} = \mathbf{Pa}^{G_{out}}_{T_j}$, and $\mathbf{V}_{X_i} = \mathbf{Pa}^{G_{out}}_{X_i}$. We also denote $\mathbf{V}^{-T_j}_{T_\ell} = \mathbf{V}_{T_\ell} - \{T_j\}$ and similarly for $\mathbf{V}^{-T_j}_{X_i}$. To simplify the presentation, we also assume that $\mathbf{U}_j \cap \mathbf{V}_{T_j} = \emptyset$.

In addition, we use the notation

$$E_{P(\cdot\mid\mathbf{u}_j)}[D(P(Y \mid \mathbf{Z}, \mathbf{u}_j) \| P(Y \mid \mathbf{Z}, t_j))]$$
$$= \sum_{\mathbf{Z}} P(\mathbf{Z} \mid \mathbf{u}_j) D(P(Y \mid \mathbf{Z}, \mathbf{u}_j) \| P(Y \mid \mathbf{Z}, t_j))$$
$$= E_{P(\cdot\mid\mathbf{u}_j)}[\log \frac{P(Y \mid \mathbf{Z}, \mathbf{u}_j)}{P(Y \mid \mathbf{Z}, t_j)}]$$

where $Y$ and $\mathbf{Z}$ are variables (or sets of variables) and $P(\cdot \mid \mathbf{u}_j)$ is the joint distribution over all variables given the specific value of $\mathbf{U}_j$. Note that this terms implies averaging over all values of $Y$ and $\mathbf{Z}$ using the conditional distribution. In particular, if $Y$ or $\mathbf{Z}$ intersects with $\mathbf{U}_j$, then only the values consistent with $\mathbf{u}_j$ have positive weights in this averaging. Also note that if $\mathbf{Z}$ is empty, then this is term reduces to the standard KL divergence between $P(Y \mid \mathbf{u}_j)$ and $P(Y \mid t_j)$.

The main result of this section is as follows.

**Theorem 7.1:** *Assume that $P(\mathbf{X})$, $G_{in}$, $G_{out}$, and $\beta$ are given. The conditional distributions $\{P(T_j \mid \mathbf{U}_j)\}$ are a stationary point of $\mathcal{L}^{(1)} = \mathcal{I}^{G_{in}} - \beta\mathcal{I}^{G_{out}}$ if an only if*

$$P(t_j \mid \mathbf{u}_j) = \frac{P(t_j)}{Z_{T_j}(\mathbf{u}_j, \beta)} e^{-\beta d(t_j, \mathbf{u}_j)} \quad (6)$$

*where $Z_{T_j}(\mathbf{u}_j, \beta)$ is a normalization term, and $d(t_j, \mathbf{u}_j)$ is given by*

$$\sum_{i:T_j \in \mathbf{V}_{X_i}} E_{P(\cdot\mid\mathbf{u}_j)}[D(P(X_i \mid \mathbf{V}^{-T_j}_{X_i}, \mathbf{u}_j) \| P(X_i \mid \mathbf{V}^{-T_j}_{X_i}, t_j))]$$
$$+ \sum_{\ell:T_j \in \mathbf{V}_{T_\ell}} E_{P(\cdot\mid\mathbf{u}_j)}[D(P(T_\ell \mid \mathbf{V}^{-T_j}_{T_\ell}, \mathbf{u}_j) \| P(T_\ell \mid \mathbf{V}^{-T_j}_{T_\ell}, t_j))]$$
$$+ E_{P(\cdot\mid\mathbf{u}_j)}[D(P(\mathbf{V}_{T_j} \mid \mathbf{u}_j) \| P(\mathbf{V}_{T_j} \mid t_j))].$$

*where all probabilities in this term are derived from the definition of the model in Eq. (1).*[3]

---

[3]This can be done by standard variable elimination procedures.

See Appendix A for a proof outline.

The essence of this theorem is that it defines $P(t_j \mid \mathbf{u}_j)$ in terms of the *distortion* $d(t_j, \mathbf{u}_j)$. This distortion measures how close are the conditional distribution in which $t_j$ is involved into these where we replace $t_j$ with $\mathbf{u}_j$. In other words, we can understand this as measuring how well $t_j$ performs as a "representative" of the particular assignment $\mathbf{u}_j$. The conditional distribution $P(T_j \mid \mathbf{u}_j)$ depends on the differences in the distortion for different values of $T_j$.

The theorem also allows us to understand the role of $\beta$. When $\beta$ is small, the conditional distribution is diffused, since $\beta$ reduces the differences between the distortions for different values of $T_j$. On the other hand, when $\beta$ is large, the exponential term acts as a "softmax" gate, and most of the conditional probability mass will be assigned to the value $t_j$ with the smallest distortion. This behavior matches the intuition that when $\beta$ is small, most of the emphasis is on compressing the input variables $\mathbf{U}_j$ into $T_j$ and when $\beta$ is large, most of the emphasis is on predicting the "outputs" variables of $T_j$, as specified by $G_{out}$.

**Example 7.2:** To see a concrete example, we reconsider the parallel bottleneck of Example 6.1. Applying the theorem to $\mathcal{L}^{(1)}$ of Eq. (4), we get that the distortion term for $T_1$ is

$$d(t_1, a) = E_{P(\cdot\mid a)}[D(P(B \mid a, T_2) \| P(B \mid t_1, T_2))]$$

This term corresponds to the information of $B$ and $T_1, T_2$. We see that $P(t_1 \mid a)$ increases when the predictions of $B$ given $t_1$ are similar to those given $a$ (when averaging over $T_2$). The distortion for $T_2$ is defined analogously.

**Example 7.3:** Consider now the symmetric bottleneck case of $G^{(a)}_{out}$ in Example 6.2. Applying the theorem, we get that the distortion term for $T_A$ is

$$d(t_A, a) = E_{P(\cdot\mid a)}[D(P(T_B \mid a) \| P(T_B \mid t_A))] +$$
$$E_{P(\cdot\mid a)}[D(P(A \mid a) \| P(A \mid t_A))]$$

The first term is a simple KL divergence, and last term can be simplified to $-\log P(a \mid t_A) = -\log p(t_A) - \log P(a) + \log P(t_A \mid a)$. By simple arithmetic operations, and using $\gamma = \frac{\beta}{1-\beta}$, we get the set of self-consistent equations

$$P(t_A \mid a) = \frac{P(t_A)}{Z_{T_A}(a, \gamma)} e^{-\gamma D(P(T_B\mid a)\|P(T_B\mid t_A))}$$
$$P(t_B \mid b) = \frac{P(t_B)}{Z_{T_B}(b, \gamma)} e^{-\gamma D(P(T_A\mid b)\|P(T_A\mid t_B))}$$

Thus, $T_A$ attempts to make predictions as similar to these of $A$ about $T_B$, and similarly $T_B$ attempts to make predictions as similar to these of $B$ about $T_A$.

## 8 Iterative Optimization Algorithm

We now consider algorithms for constructing solutions of the variational principle.

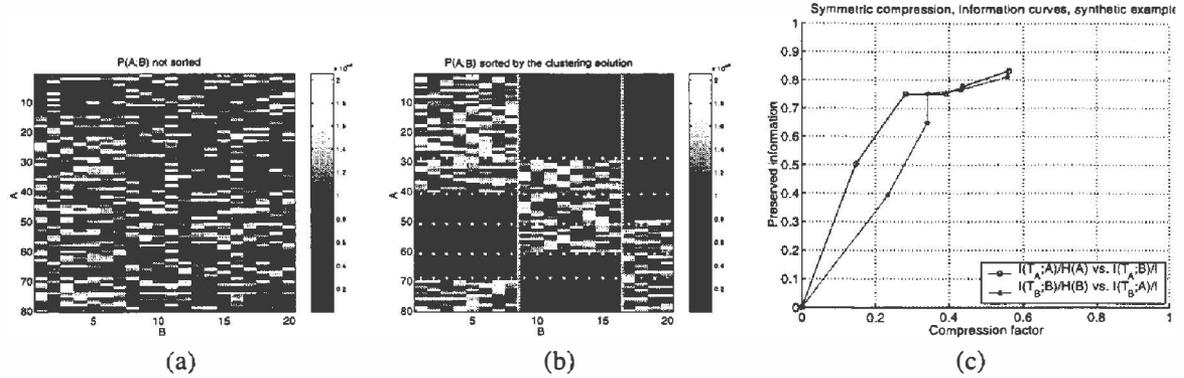

Figure 3: Application of the symmetric bottleneck on a simple synthetic example. (a) input joint distribution, (b) the same joint distribution where rows and columns were permuted to match the clustering found; dotted lines show cluster boundaries. (c) information curves showing the progression along the information tradeoff graph for increasing $\beta$. The $x$-axis is the fraction of the information about the original variable that is maintained by the compressed variable, and the $y$-axis is the fraction of the information between $A$ and $B$ that is captured by $I(T_A; B)$ or $I(T_B; A)$. Circles denote bifurcation events.

We start with the case where $\beta$ is fixed. In this case, following standard strategy in variational methods, we simply apply the self-consistent equations. More precisely, we use an iterative algorithm, that at the $m$'th iteration maintains the conditional distributions $\{P^{(m)}(T_j \mid \mathbf{U}_j) : j = 1, \ldots, k\}$. At the $m+1$'th iteration, the algorithm applies an *update step*

$$P^{(m+1)}(t_j \mid \mathbf{u}_j) \leftarrow \frac{P^{(m)}(t_j)}{Z_{T_j}^{(m+1)}(\mathbf{u}_j, \beta)} e^{-\beta d^{(m)}(t_j, \mathbf{u}_j)} \quad (7)$$

where $P(t_j)^{(m)}$ and $d^{(m)}(t_j, \mathbf{u}_j)$ are computed with respect to the conditional probabilities $\{P^{(m)}(T_j \mid \mathbf{U}_j) : j = 1, \ldots k\}$.

There are two main variants of this algorithm. In the *synchronous* variant, we apply the update step for all the conditional distributions in each iteration. That is, each conditional probability $P(T_j \mid \mathbf{U}_j)$ is updated by computing the distortion based on the conditional probabilities of the previous iterations. In the *asynchronous* variant, we choose one variable $T_j$, and perform the update only for this variable. For all $\ell \neq j$, we set $P^{(m+1)}(T_\ell \mid \mathbf{U}_\ell) = P^{(m)}(T_\ell \mid \mathbf{U}_\ell)$. The main difference between the two variants is that the update of $T_j$ in the asynchronous update incorporates the implications of the updates of the other variables.

**Theorem 8.1 :** *Asynchronous iterations of the self-consistent equations converge to a stationary point of the optimization problem.*

See Appendix A for a proof for the case $\beta < 1$. The convergence proof for the general case is more involved and will appear in the full version of this paper.

At the current stage we do not have a proof of convergence for the synchronous case, although in all our experiments, synchronous updates converge as well.

A key question is how to initialize this procedure, as different initialization points can lead to different solutions. We now describe a *deterministic annealing* like procedure [8, 13]. This procedure works by iteratively increasing the parameter $\beta$ and then adapting the solution for the previous value of $\beta$ to the new one. This allows the algorithm to "track" the changes in the solution as the system shifts its preferences from compression to prediction.[4]

Recall that when $\beta \to 0$, the optimization problem tends to make $T_j$ independent of its parents. At this point the solution consists of essentially only one cluster for each $T_j$ which is not predictive about any other variable. As we increase $\beta$, we suddenly reach a point where the values of $T_j$ diverge and show two different behaviors. This phenomena is a phase-transition of the system. Successive increases of $\beta$ will reach additional phase transitions in which additional splits of values of $T_j$ emerge. The general idea of this annealing procedure is to identify these *bifurcations* of clusters. At each step of the procedure, we maintain a set of values for each $T_j$. Initially, when $\beta = 0$, each $T_j$ has a single value. We then progressively increase $\beta$ and try to detect bifurcations. At the end of the procedure we record for each $T_j$ a bifurcating tree that traces the sequence of solutions at different values of $\beta$ (see for example Figure 4(a)).

The main technical problem is how to detect such bifurcations. We adopt the methods of Tishby *et al.* [13] to multiple variables. At each step, we take the solution from the previous step (i.e., for the previous value of $\beta$ we considered) and construct an initial problem in which we duplicate each value of each $T_j$. To define such an initial solution we need to specify the conditional probabilities of these "doubled" values given each value $\mathbf{U}_j$. Sup-pose that $t_{j,a}$ and $t_{j,b}$ are the two copies of the value $t_j$. Then we set $P^*(t_{j,a} \mid \mathbf{u}_j) = P(t_j \mid \mathbf{u}_j)\left(\frac{1}{2} + \alpha\epsilon(\mathbf{u}_j)\right)$ and $P^*(t_{j,b} \mid \mathbf{u}_j) = P(t_j \mid \mathbf{u}_j)\left(\frac{1}{2} - \alpha\epsilon(\mathbf{u}_j)\right)$ where $\epsilon(\mathbf{u}_j) \sim U[-\frac{1}{2}, \frac{1}{2}]$ is a noise term and $0 < \alpha \le 1$ is a scale parameter. Thus, each copy $t_{j,a}$ and $t_{j,b}$ is a *perturbed* version of $t_j$. If $\beta$ is high enough, this random perturbation suffices to allow the two copies of $t_j$ to diverge. If $\beta$ is too small to support such bifurcation, both perturbed versions will collapse to the same solution.

---

[4] In deterministic annealing terminology, $\frac{1}{\beta}$ is the "temperature" of the system, and thus increasing $\beta$ corresponds to "cooling" the system.

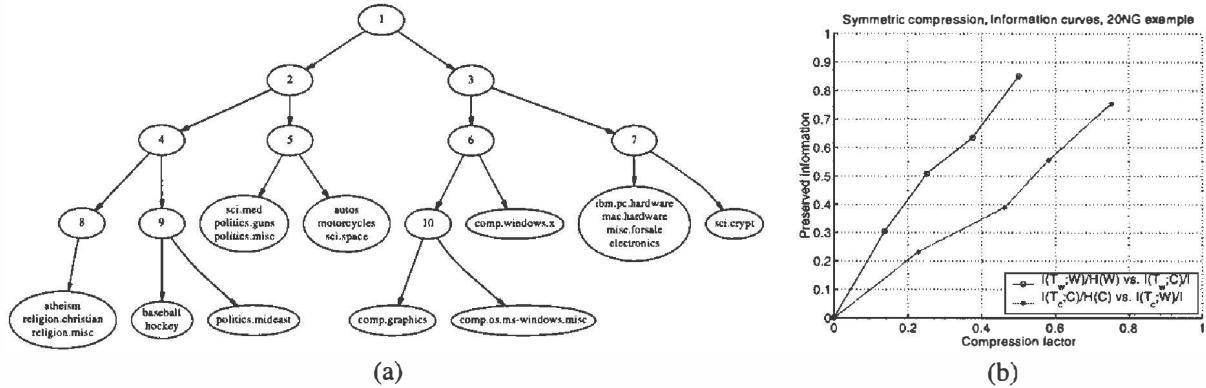

Figure 4: Application of the symmetric bottleneck to the 20 newsgroup data set with 300 informative words. (a) The learned cluster hierarchy of categories. (b) information curves showing the progression along the information tradeoff graph. Note that with 16 word clusters we preserve most of the information present in the data.

After constructing this initial point, we iteratively perform the update equations of (7) until convergence. If at the convergence point the behavior of $t_{j,a}$ and $t_{j,b}$ is identical, then we declare that the value $t_j$ has not split. On the other hand, if the distribution $P(t_{j,a} \mid \mathbf{u}_j)$ is sufficiently different from $P(t_{j,b} \mid \mathbf{u}_j)$ for some values of $\mathbf{U}_j$, then we declare that the value $t_j$ has split, and incorporate $t_{j,a}$ and $t_{j,b}$ into the bifurcation we construct for $T_j$. Finally, we increase $\beta$ and repeat the whole process.

We stress that this annealing procedure is heuristic in nature. We do not have formal guarantees that it will find the global optima. Nonetheless, it has the distinct advantage in that charts the behavior of the system at different values of $\beta$.

An alternative and simpler approach that proved useful for the original IB formulation employed agglomerative clustering techniques to find a bifurcating tree in a bottom-up fashion [9]. Such an approach can also applied to the multivariate case and will be presented elsewhere.

## 9　Examples

To illustrate the ideas described above, we now examine few applications of symmetric and parallel versions of the bottleneck.

As a simple synthetic example we produced a joint probability matrix $P(A, B)$ (see Figure 3(a)) where $|A| = 80$ and $|B| = 20$. Using the symmetric compression ($\mathcal{L}_a^{(2)}$ of Example 6.2) we find 6 natural clusters for $A$ and 3 natural clusters for $B$. Sorting the joint probability matrix by this solution illustrates this structure (Figure 3(b)). It is also interesting to see the fraction of information preserved by our clusters. One way of presenting these results is by considering the fraction of information preserved by $T_A$ about $B$, and analogously, the fraction of information preserved in $T_B$ about $A$. This amount of preserved information should be plotted with respect to the compression factor, i.e., how compact is the new clusters representation. This is given of course by $I(T_A; A)$ and $I(T_B; B)$ respectively. In Figure 3(c) we present these two information curves. In both curves we see that splitting the current clusters set increase the amount of information preserved about the relevant variable, and simultaneously reduces the compression (since more clusters induces less compression). In the first split we find 2 clusters in $T_A$ and 2 in $T_B$. The second split results with 4 clusters in $T_A$ and 3 in $T_B$. The next split finds 6 clusters in $T_A$ and leaves $T_B$ with 3 clusters. This is indeed the "real" structure of this data. Interestingly, due to this last split, the information $T_B$ preserve about $A$ is increased, though there was no split in $T_B$. The reason, of course, is that $T_B$ predicts $A$ *through* $T_A$, thus the split in $T_A$ increases $I(T_B; A)$. On the other hand, since there was no split (at this step) in $T_B$, $I(T_A; B)$ remains unchanged. The next splits are practically overfitting effects and accordingly there is no real information gain due to these splits.

As a more realistic example we used the standard 20 newsgroups corpus. This natural language corpus contains about 20,000 articles evenly distributed among 20 USENET discussion groups [5] and has been employed for evaluating text classification techniques (e.g., [12]). Many of these groups have similar topics. Five groups discuss different issues concerning computers, three groups discuss religion issues, etc. Thus, there is an inherent hierarchy among these groups.

To model this domain in our setting, we introduce two random variables. We let $W$ denote words, and $C$ denote a category (i.e., a newsgroup). The joint probability $P(w, c)$ is the probability that a random word-position (e.g., word 218 in document 1255) in this collection is equal to $w$ and at the same time the category of the document is $c$. To obtain such a joint distribution we performed several pre-processing steps: We removed file headers, transformed all words to lower case, and removed stop words and words containing digits or non-alphanumeric characters. We then sorted all words by the contribution to the mutual information about the category variable. More formally, we sorted all words by $I(w) \equiv P(w) \sum_{c \in C} P(c \mid w) \log \frac{P(c|w)}{P(c)}$, and used the subset of the top 300 most "informative" words. After re-normalization, we had a joint probability matrix with $|W| = 300$ and $|C| = 20$.

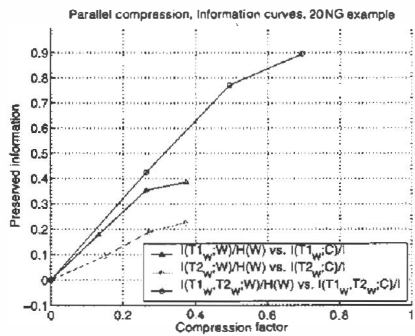

Figure 5: Information curves for parallel compression of the 20 newsgroup data set showing the progression along the information tradeoff graph.

We first used the symmetric bottleneck algorithm to cluster both dimensions of this matrix into two sets of clusters: clusters of words, $T_w$, and clusters of categories, $T_c$. The hierarchy found in $T_c$, shown in Figure 4(a) is in high agreement with the natural hierarchy one would construct. Additionally, each of the word clusters is in high correlation with one of these category clusters. For example, for the second word cluster, the 5 most probable words (i.e. the 5 words that maximize $P(t_{w_2} \mid W)$), were 'islamic', 'religious', 'homosexual', 'peace' and 'religion'. Accordingly $P(T_c \mid t_{w_2})$ was maximized for the "religion" cluster in $T_c$ (left cluster in Figure 4(a)).

As already explained, the general mapping scheme we use is a "soft" one. That is, each object could be assigned to each cluster with some normalized probability. The clustering of $C$ into $T_c$ was typically "hard" (for every $c \in C$, $P(T_c \mid c)$ was approximately 1 for one cluster and 0 for the others). However, the clustering of $W$ into $T_w$ utilized the "soft" aspect of the clustering to deal with words that are relevant to several category clusters. Thus, some of the words were assigned to more than one cluster. For example, the word 'Clinton' was assigned to two different word clusters dealing with politics. The word 'sexual' was assigned to the same two clusters, and also (with lower probability) to a cluster of words dealing with religious issues.

For the same data we used also the parallel compression ($\mathcal{L}^{(1)}$ of Example 6.1). In this case we have two compression variables, $T_{1,w}, T_{2,w}$, that try simultaneously (and independently), to cluster the set of words $W$ in a way that will preserve the information about the category variable $C$, as high as possible. In Figure 5, we present the information curves for these two cluster sets. Clearly, using the combination of the compression variables is much more informative than using each one of them independently. For example, after the second split, $|T_{1,w}| = |T_{2,w}| = 4$, and $I(T_{1,w}; C)$ and $I(T_{2,w}; C)$ preserve 38% and 20% of the original information $I(W; C)$, respectively. On the other hand, at the same stage, $I(T_{1,w}, T_{2,w}; C)$ preserves almost 80% of $I(W; C)$. Thus, only 8 word-clusters are enough to preserve most of the information about the category variable.

## 10 Discussion

We presented a novel general framework for data analysis. This new framework provides a natural generalization of the information bottleneck method. Moreover, as we have shown, it immediately suggests new bottleneck like constructions, and provides generic tools to implement them.

Many connections with other data analysis methods should be explored. The general structure of the iterative procedure is reminiscent of EM and k-means procedures. Other connections are, for example, to dimensionality reduction techniques, such as ICA [2]. The parallel bottleneck construction provides an ICA-like decomposition with an important distinction. In contrast to ICA, it is aimed at preserving information about specific aspects of the data, defined by the user.

The suggested framework allows us to extract structure from data in numerous ways. In our examples, we explored only few relatively simple cases, but clearly this is the tip of the iceberg. We are currently working on several additional applications under this framework. These include analysis of gene expression data, neural coding and DNA sequence analysis, document clustering, and computational linguistic applications.

### Acknowledgements

This work was supported in part by the Israel Science Foundation (ISF), the Israeli Ministry of Science, and by the US-Israel Bi-national Science Foundation (BSF). N. Slonim was also supported by an Eshkol fellowship. N. Friedman was also supported by an Alon fellowship and the Harry & Abe Sherman Senior Lectureship in Computer Science. Experiments reported here were run on equipment funded by an ISF Basic Equipment Grant.

## References

[1] L. D. Baker and A. K. McCallum. Distributional clustering of words for text classification. In *ACM SIGIR 98*. 1998.

[2] A.J. Bell & T.J. Sejnowski. An information-maximization approach to blind separation and blind deconvolution. Neur. Comp. 7, 1129-1159, 1995.

[3] T. M. Cover and J. A. Thomas. *Elements of Information Theory*. John Wiley & Sons, New York, 1991.

[4] T. Hofmann. Probabilistic latent semantic indexing. In *ACM SIGIR 99*, pages 50–57. 1999.

[5] K. Lang. Learning to filter netnews. In *12th Int. Conf. on Machine Learning*, pages 331-339. 1995.

[6] J. Pearl. *Probabilistic Reasoning in Intelligent Systems*. Morgan Kauffman, San Francisco, 1988.

[7] F.C. Pereira, N. Tishby, and L. Lee. Distributional clustering of English words. In *30th Annual Meeting of the Association for Computational Linguistics*, pages 183-190. 1993.

[8] K. Rose. Deterministic annealing for clustering, compression, classification, regression, and related optimization problems. *Proceedings of the IEEE*, 86:2210-2239, 1998.

[9] N. Slonim & N. Tishby. *Agglomerative Information Bottleneck*. in Advances in Neural Information Processing Systems (NIPS) 12, pp. 617-623, 1999.

[10] N. Slonim, R. Somerville, N. Tishby, and O. Lahav. Objective spectral classification of galaxies using the information bottleneck method. in "Monthly Notices of the Royal Astronomical Society", MNRAS, 323, 270, 2001.

[11] N. Slonim and N. Tishby. Document clustering using word clusters via the information bottleneck method. In *ACM SIGIR 2000*, pages 208-215. 2000.

[12] N. Slonim and N. Tishby. The power of word clusters for text classification. In *23rd European Colloquium on Information Retrieval Research*. 2001.

[13] N. Tishby, F. Pereira, and W. Bialek. The information bottleneck method. In *Proc. 37th Allerton Conference on Communication and Computation*. 1999.

[14] N. Tishby and N. Slonim. Data clustering by Markovian relaxation and the information bottleneck method. In *Neural Information Processing Systems (NIPS-00)*. 2000.

## A  Proofs

We now sketch the proofs of the two main theorems. We start with Theorem 7.1.

**Proof:** The basic idea is to find stationary points of $\mathcal{L}^{(1)}$ subject to the normalization constraints. Thus, we add Lagrange multipliers and get the Lagrangian
$$J = \mathcal{I}^{G_{in}} - \beta \mathcal{I}^{G_{out}} - \sum_j \sum_{\mathbf{u}_j} \lambda_{\mathbf{u}_j} \sum_{t_j} P(t_j \mid \mathbf{u}_j).$$
To differentiate $J$ we use the following lemma.

**Lemma A.1:** $\frac{\partial I(Y;Z)}{\partial P(t_j|\mathbf{u}_j)} = P(\mathbf{u}_j)(E_{P(\cdot|t_j,\mathbf{u}_j)}[\log \frac{P(Y|Z)}{P(Y)}] - 1)$.

We now can differentiate each mutual information term that appears in $J$. Note that we can ignore terms that do not depend on the value of $T_j$, since these will be absorbed by the normalization constant. Thus, a term of the form $E_{P(\cdot|t_j,\mathbf{u}_j)}[\log P(Y \mid Z)]$ where $T_j \notin \mathbf{Z} \cup \{Y\}$ can be ignored. Collecting terms that do refer to $t_j$, equating to 0, and dividing by $P(\mathbf{u}_j)$ we get the following equation.

$$\begin{aligned}\log P(t_j \mid \mathbf{u}_j) &= \log P(t_j) \\ &+ \beta \sum_{i:T_j \in \mathbf{V}_{X_i}} E_{P(\cdot|t_j,\mathbf{u}_j)}[\log P(X_i \mid \mathbf{V}_{X_i})] \\ &+ \beta \sum_{\ell:T_j \in \mathbf{V}_{T_\ell}} E_{P(\cdot|t_j,\mathbf{u}_j)}[\log P(T_\ell \mid \mathbf{V}_{T_\ell})] \\ &+ \beta E_{P(\cdot|t_j,\mathbf{u}_j)}[\log \frac{P(t_j \mid \mathbf{V}_{T_j})}{P(t_j)}] \\ &+ c_j(\mathbf{u}_j) \end{aligned} \quad (8)$$

where $c_j(\mathbf{u}_j)$ is a term that does not depend on $t_j$.

To get the desired form of the self-consistent equations, we apply several manipulations. First, we can write $E_{P(\cdot|t_j,\mathbf{u}_j)}[\log P(X_i \mid \mathbf{V}_{X_i})] = E_{P(\cdot|\mathbf{u}_j)}[\log P(X_i \mid \mathbf{V}_{X_i}^{-T_j}, t_j)]$ since all the variables in $\mathbf{V}_{X_i}^{-T_j}$ are independent of $T_j$ given $\mathbf{U}_j$. Second, we can transform the latter term into a KL divergence by extracting the term $-E_{P(\cdot|\mathbf{u}_j)}[\log P(X_i \mid \mathbf{V}_{X_i}^{-T_j}, \mathbf{u}_j)]$ from $c_j(\mathbf{u}_j)$. Similar transformation applies to the terms that deal with $T_\ell$. Third, we use Bayes rule to rewrite $E_{P(\cdot|t_j,\mathbf{u}_j)}[\log \frac{P(t_j|\mathbf{V}_{T_j})}{P(t_j)}] = E_{P(\cdot|t_j,\mathbf{u}_j)}[\log \frac{P(\mathbf{V}_{T_j}|t_j)}{P(\mathbf{V}_{T_j})}]$. Since $P(\mathbf{V}_{T_j})$ does not involve $t_j$ we can ignore it. To get a KL divergence term, we subtract the term $E_{P(\cdot|t_j,\mathbf{u}_j)}[\log \mathbf{V}_{T_j} \mid \mathbf{u}_j]$. Finally, we apply the normalization constraints for each distribution $P(T_j \mid \mathbf{u}_j)$ to get the desired equations. ∎

We now turn to the proof of Theorem 8.1.

**Proof:** To prove convergence it suffices to prove that unless we are at a stationary point, each application of the assignment Eq. (7) reduces the Lagrangian $\mathcal{L}^{(1)}[P] = \mathcal{I}^{G_{in}} - \beta \mathcal{I}^{G_{out}}$. Recall, that we require that $P \models G_{in}$. Also, recall that when $\beta < 1$, minimizing this Lagrangian is equivalent to minimizing the Lagrangian $\mathcal{L}^{(2)}[P] = \mathcal{I}^{G_{in}} + \frac{\beta}{1-\beta} D(P \| G_{out})$.

To show convergence, we will introduce an auxiliary Lagrangian: $\mathcal{F}[P, R, Q] = D(P \| R) + \frac{\beta}{1-\beta} D(P \| Q)$ subject to the constraints that $P \models G_{in}$, $Q \models G_{out}$, and $R \models G_\emptyset$, where $G_\emptyset$ is the DAG without edges. It is easy to see that $\mathcal{L}^{(2)}$ and $\mathcal{F}$ coincide when $Q = \prod_i P(X_i \mid \mathbf{V}_{X_i}) \prod_j P(T_j \mid \mathbf{V}_{T_j})$ and $R = \prod_i P(X_i) \prod_j P(T_j)$. That is, when $Q$ and $R$ are the KL-projections of $P$ onto the space of distributions consistent with $G_{out}$ and $G_{in}$.

Using properties of KL-projections, we get.

**Lemma A.2:** *For any choice of $P$, $R$ and $Q$ that are consistent with $G_{in}$, $G_{out}$ and $G_\emptyset$, respectively, $\mathcal{L}^{(2)}[P] \leq \mathcal{F}[P, Q, R]$, with equality if and only if $Q$ and $R$ are the projections of $P$ onto $G_{out}$ and $G_\emptyset$, respectively.*

Assume that $Q$ and $R$ are fixed, and suppose we want to modify $P(T_j \mid \mathbf{u}_j)$ to minimize $\mathcal{F}$. Taking derivatives of $\mathcal{F}$ with respect to $P(t_j \mid \mathbf{u}_j)$ and equating to 0, we get the following self consistent equations:

$$P(t_j \mid \mathbf{u}_j) = \frac{R(t_j)}{Z(\beta, \mathbf{u}_j)} e^{\beta \left( E_{P(\cdot|t_j,\mathbf{u}_j)}[\log Q] - \log R(t_j) \right)}$$

It is easy to verify that the right hand side of this equation does not depend on $P(t_j \mid \mathbf{u}_j)$. Thus, the assignment

$$P'(t_j \mid \mathbf{u}_j) \leftarrow \frac{R(t_j)}{Z(\beta, \mathbf{u}_j)} e^{\beta \left( E_{P(\cdot|t_j,\mathbf{u}_j)}[\log Q] - \log R(t_j) \right)}$$

and $P'(t_\ell \mid \mathbf{u}_\ell) \leftarrow P(t_\ell \mid \mathbf{u}_\ell)$ for $\ell \neq j$ results in a distribution $P'$ such that $\mathcal{F}[P', Q, R]$ is a stationary point with respect to changes in $P(T_j \mid \mathbf{U}_j)$. Moreover, it is easy to verify that the second derivative of $\mathcal{F}$ with respect to $P(t_j \mid \mathbf{u}_j)$ is positive, and thus this point is a local minima. We conclude that $\mathcal{F}[P', Q, R] \leq \mathcal{F}[P, Q, R]$, with equality if and only if $P(T_j \mid \mathbf{U}_j)$ minimizes $\mathcal{F}$ (with respect to fixed choice of $Q$, $R$, and $P(T_\ell \mid \mathbf{U}_\ell)$ for $\ell \neq j$).

We now put all these together. Suppose that $Q$ and $R$ are the projections of $P$ on $G_{out}$ and $G_\emptyset$, respectively. Then,

$$\mathcal{L}^{(2)}[P'] \leq \mathcal{F}[P', Q, R] \leq \mathcal{F}[P, Q, R] = \mathcal{L}^{(2)}[P].$$

Moreover, we have equality only if $P' = P$. This shows that an update step reduces the value of the Lagrangian. The only situation where the value remains the same is when the self consistent equation for $P(T_j \mid \mathbf{U}_j)$ is satisfied.

The only remaining issue is to show that this iteration is equivalent to the asynchronous iteration of Eq. (7) when $Q$ and $R$ are the projections of $P$ on $G_{out}$ and $G_\emptyset$, respectively. This is can be easily verified by comparing to Eq. (8). ∎